\documentclass[10pt,twocolumn,letterpaper]{article}

\usepackage{cvpr}
\usepackage{times}
\usepackage{epsfig}
\usepackage{graphicx}
\usepackage{amsmath}
\usepackage{amssymb}
\usepackage{caption}
\usepackage{subcaption}

\usepackage[pagebackref=true,breaklinks=true,letterpaper=true,colorlinks,bookmarks=false]{hyperref}

\cvprfinalcopy 

\begin{document}

\title{A Joint Speaker-Listener-Reinforcer Model for Referring Expressions}

\author{Licheng Yu, Hao Tan, Mohit Bansal, Tamara L. Berg\\
Department of Computer Science\\
University of North Carolina at Chapel Hill\\
{\tt\small \{licheng, airsplay, mbansal, tlberg\}@cs.unc.edu}\\
}

\maketitle

\begin{abstract}
Referring expressions are natural language constructions used to identify particular objects within a scene.
In this paper, we propose a unified framework for the tasks of referring expression comprehension and generation.
Our model is composed of three modules: speaker, listener, and reinforcer. The speaker generates referring expressions, the listener comprehends referring expressions, and the reinforcer introduces a reward function to guide sampling of more discriminative expressions.
The listener-speaker modules are trained jointly in an end-to-end learning framework, allowing the modules to  be aware of one another during learning while also benefiting from the discriminative reinforcer's feedback. 
We demonstrate that this unified framework and training achieves state-of-the-art results for both comprehension and generation on three referring expression datasets. Project and demo page: \href{https://vision.cs.unc.edu/refer}{https://vision.cs.unc.edu/refer}.
\end{abstract}

\vspace{-.2cm}
\section{Introduction}
\vspace{-.1cm}

People often use referring expressions in their everyday discourse to unambiguously identify or indicate particular objects within their physical environment. For example, one might point out a person in the crowd by referring to them as ``the man in the blue shirt'' or you might ask someone to ``pass me the red pen on the table.'' In both of these examples, we have a pragmatic interaction between two people (or between a person and an intelligent agent such as a robot). First, we have a speaker who must generate an expression given a target object and its surrounding world context. Second, we have a listener who must interpret and comprehend the expression and map it to an object in the environment.  Therefore, in this paper we propose an end-to-end trained listener-speaker framework that models these behaviors jointly.  

In addition to the listener and speaker, we also introduce a new reinforcer module that learns a discriminative reward model to help generate less ambiguous expressions (expressions that apply to the target object but not to other objects in the image). This goal corresponds to the Gricean Maxim~\cite{Grice75} of manner, where one tries to be as clear, brief, and orderly as possible while avoiding obscurity and ambiguity. Avoiding ambiguity is important because the generated expression should be easily and uniquely mapped to the target object. For example, if there were two pens on the table one ``long and red'' and the other ``short and red'', asking for the ``red pen'' would be ambiguous while asking for the ``long pen'' would be better. The reinforcer module is incorporated using reinforcement learning, inspired by behavioral psychology that says that agents operating in an environment should take actions that maximize their expected cumulative reward. In our case, the reward takes the form of a discriminative classifier trained to reward the speaker for generating less ambiguous expressions.

\begin{figure*}[t]
\centering
\includegraphics[width=0.9\textwidth]{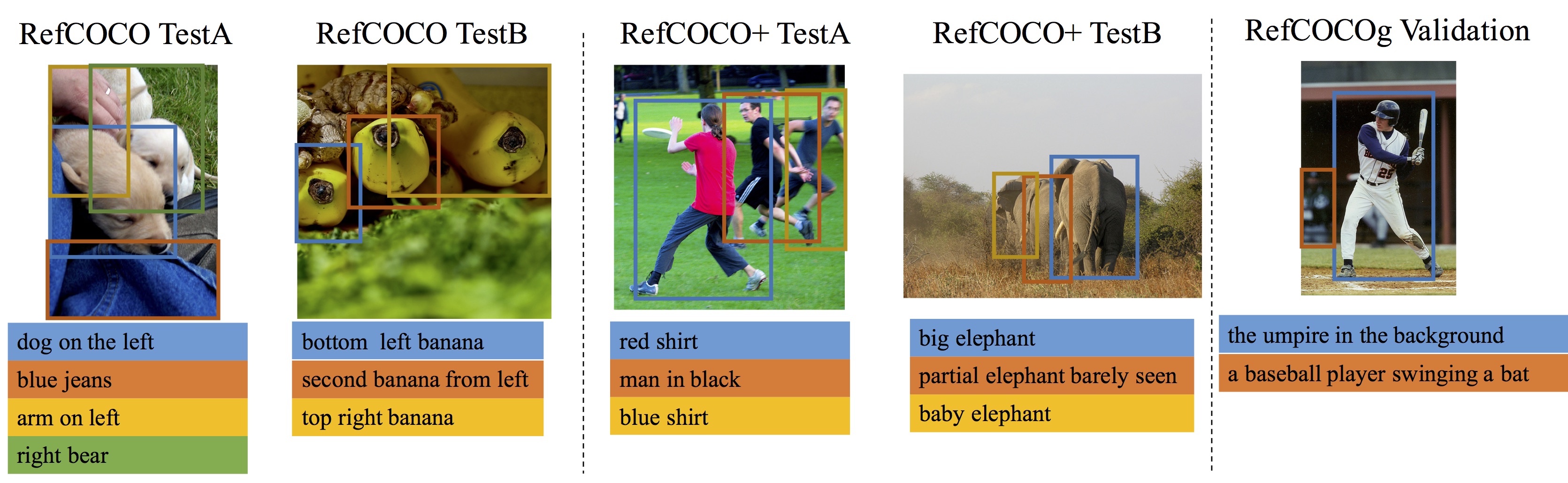}
\caption{Joint generation examples using our full model with ``+rerank'' on three datasets. Each sentence shows the generated expression for one of the depicted objects (color coded to indicate correspondence).}
\label{fig:example_generation}
\end{figure*}

\begin{figure*}[t]
\centering
\includegraphics[width=0.9\textwidth]{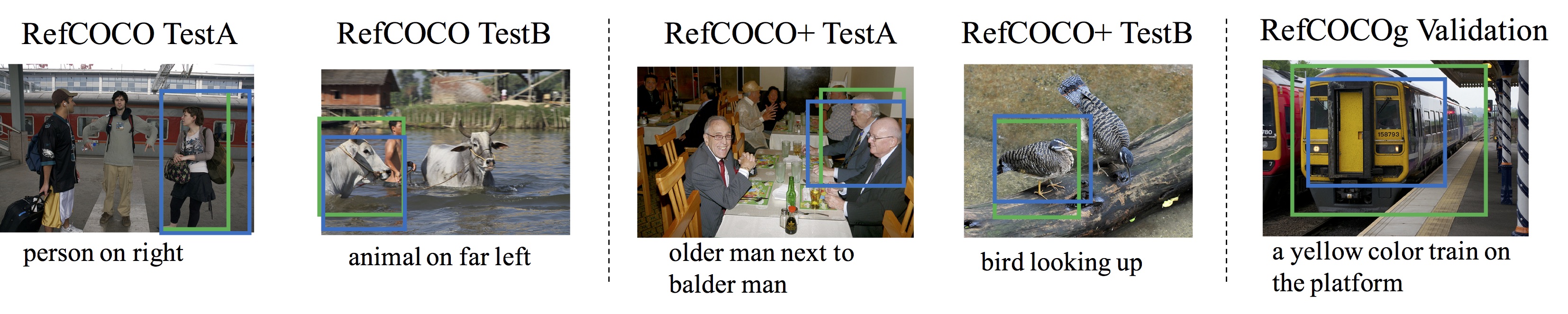}
\caption{Example comprehension results using our full model on three datasets. Green box shows the ground-truth region and blue box shows our correct comprehension based on the detected regions.}
\label{fig:example_comprehension}
\end{figure*}

Within the realm of referring expressions, there are two tasks that can be computationally modeled, mimicking the listener and speaker roles. 
Referring Expression Generation (speaker) requires an algorithm to generate a referring expression for a given target object in a visual scene, as shown in Fig.~\ref{fig:example_generation}. 
Referring Expression Comprehension (listener) requires an algorithm to localize the object/region in an image described by a given referring expression, as shown in Fig.~\ref{fig:example_comprehension}. 

The Referring Expression Generation (REG) task  has been studied since the 1970s~\cite{winograd1972understanding}.
Many of the early works in this space focused on relatively limited datasets, using synthesized images of objects in  artificial scenes or limited sets of real-world objects in simplified environments~\cite{mitchell2013generating, golland2010game, krahmer2012computational}. 
Recently, the research  focus has shifted to more complex natural image datasets and has expanded to include the Referring Expression Comprehension task~\cite{kazemzadeh2014referitgame, mao2015generation, yu2016refer} as well as to real-world interactions with robotics~\cite{fang2015embodied, eldoninterpreting}.
One reason this has become feasible is 
that several large-scale REG datasets have been collected at a scale where deep learning models can be applied. 
Kazemzadeh et al~\cite{kazemzadeh2014referitgame} introduced the first large-scale REG dataset on 20k natural images from the ImageClef dataset~\cite{grubinger2006iapr} using an interactive game,
later expanded to images from the MSCOCO dataset~\cite{lin2014microsoft} in Yu et al~\cite{yu2016refer}.
In addition, Mao et al~\cite{mao2015generation} collected Google's REG dataset, also based on MSCOCO images, but in a non-interactive setting, resulting in more complex lengthy expressions. In this paper, we focus our evaluations on the three recent datasets collected on MSCOCO images~\cite{yu2016refer, mao2015generation}. 

Recent neural approaches to the referring expression generation and comprehension tasks can be roughly split into two types.
The first type uses a CNN-LSTM encoder-decoder generative model~\cite{vinyals2015show} to generate (decode) sentences given the encoded target object.
With careful design of the visual representation of target object, this model can generate unambiguous expressions~\cite{mao2015generation, yu2016refer}.
Here, the CNN-LSTM models $P(r|o)$, where $r$ is the referring expression and $o$ is the target object, which can be easily converted to $P(o|r)$ via Bayes' rule and used to address the comprehension task~\cite{hu2015natural, mao2015generation, yu2016refer, nagaraja16refexp} by selecting the $o$ with the largest posterior probability.
The second type of approach uses a joint-embedding model that projects both a visual representation of the target object and a semantic representation of the expression into a common space and learns a distance metric. Generation and comprehension can be performed by embedding a target object (or expression) into the embedding space and retrieving the closest expression (or object) in this space.
This type of approach typically achieves better comprehension performance than the CNN-LSTM model as in~\cite{rohrbach2015grounding, wang2015learning}, but previously was only applied to the referring expression comprehension task. Recent work~\cite{andreas2016reasoning} has also used both an encoder-decoder model (speaker) and an embedding model (listener) for referring expression generation in abstract images, where the offline listener reranks the speaker's output.

In this paper, we propose a unified model that \emph{jointly} learns both the CNN-LSTM speaker and embedding-based listener models, for both the generation and comprehension tasks. Additionally, we add a discriminative reward-based \emph{reinforcer} to guide the sampling of more discriminative expressions and further improve our final system. Instead of working independently, we let the speaker, listener, and reinforcer interact with each other, resulting in improved performance on both generation and comprehension tasks. Results evaluated on three standard, large-scale datasets verify that our proposed listener-speaker-reinforcer model significantly outperforms the state-of-the-art on both the comprehension task (Tables~\ref{table:comprehension_speaker} and~\ref{table:comprehension_listener}) and the generation task (evaluated using  human judgements in Table ~\ref{table:human_eval}, and automatic metrics in Table~\ref{table:metric}).

\vspace{-.2cm}
\section{Related work}
\label{sec:relatedwork}
\vspace{-.1cm}

Recent years have witnessed a rise in multimodal research related to vision and language. Given the individual success in each area, and the need for models with more advanced cognition capabilities, several tasks have emerged as evaluation applications, including image captioning, visual question answering, and referring expression generation/comprehension.  

\smallskip
\noindent{\bf Image Captioning:}
The aim of image captioning is to generate a sentence describing the general content of an image.
Most recent approaches use deep learning to address this problem.
Perhaps the most common architecture is a CNN-LSTM model~\cite{vinyals2015show}, which generates a sentence conditioned on visual information from the image.
One paper related to our work is gLSTM~\cite{jia2015guiding} which uses CCA semantics to guide the caption generation.
A further step beyond image captioning is to locate the regions being described in captions~\cite{wang2016phrase, plummer2015flickr30k, wang2015learning}.
The Visual Genome~\cite{krishna2016visual} collected captions for dense regions in an image that have been used for dense-captioning tasks~\cite{johnson2015densecap}.
While there have been many approaches achieving quite good results on captioning, one challenge has been that the caption that a system should output for an image is very task dependent.
Therefore, there has been a movement toward more focused tasks, such as visual question answering~\cite{Stanislaw2015VQA, yu2015visual}, and referring expression generation and comprehension which involve specific regions/objects within an image (discussed below).

\smallskip
\noindent{\bf Referring Expression Datasets:}
REG has been studied for many years~\cite{winograd1972understanding, krahmer2012computational, mitchell2013generating} in linguistics and natural language processing, but mainly focused on small or artificial datasets.
In 2014, Kazemzadeh et al~\cite{kazemzadeh2014referitgame} introduced the first large-scale dataset RefCLEF using 20,000 real-world natural images~\cite{grubinger2006iapr}.
This dataset was collected in a two-player game, where the first player writes a referring expression given an indicated target object. The second player is shown only the image and expression and has to click on the correct object described by the expression.
If the click lies within the target object region, both sides get points and their roles switch.
Using the same game interace, the authors further collected RefCOCO and RefCOCO+ datasets on MSCOCO images~\cite{yu2016refer}.
The RefCOCO and RefCOCO+ datasets each contain 50,000 referred objects with 3 referring expressions on average. The main difference between RefCOCO and RefCOCO+ is that in RefCOCO+, players were forbidden from using absolute location words, e.g. left dog, therefore focusing the referring expression to purely appearance-based descriptions.
In addition, Mao et al~\cite{mao2015generation} also collected a referring expression dataset - RefCOCOg, using MSCOCO images, but in a non-interactive framework. These expressions are more similar to the MSCOCO captions in that they are longer and more complex as their was no time constraint in the non-interactive data collection setting. This dataset, RefCOCOg, has 96,654 objects with 1.3 expressions per object on average.

\begin{figure*}[t]
\centering
\vspace{-12pt}
\includegraphics[width=0.70\textwidth]{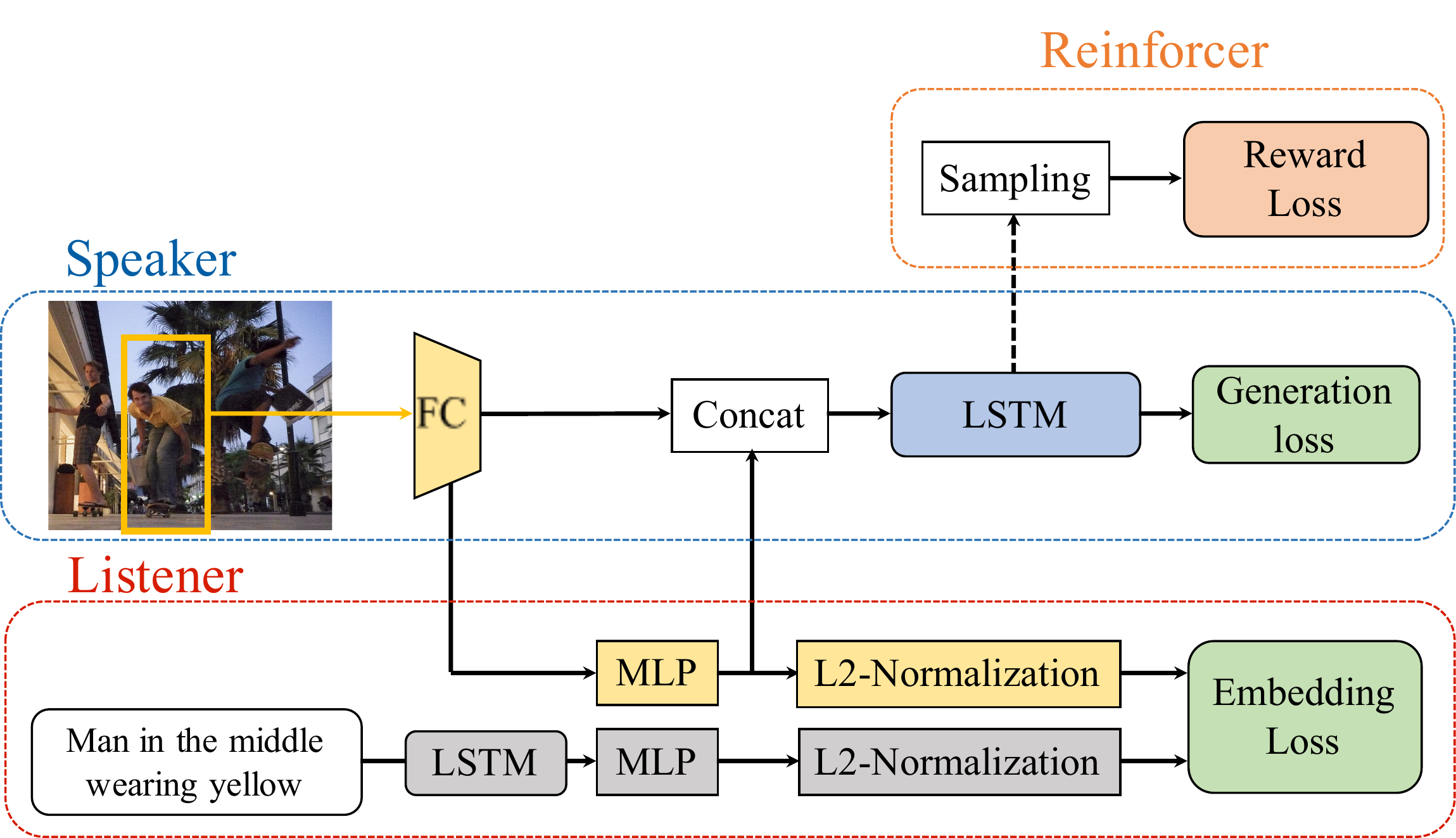}
\caption{Framework: The Speaker is a CNN-LSTM model, which  generates a referring expression for the target object. The Listener is a joint-embedding model learned to minimize the distance between paired object and expression representations. In addition, a Reinforcer module  helps improve the speaker by sampling more discriminative (less ambiguous) expressions for training.  The model is jointly trained with 3 loss functions -- generation loss, embedding loss, and reward loss, thereby improving performance on both the comprehension and generation tasks.
}
\label{fig:model}
\end{figure*}

\smallskip
\noindent{\bf Referring Expression Comprehension and Generation:}
Referring expressions are associated with two tasks, comprehension and generation.
The comprehension task requires a system to select the region being described by a given referring expression.
To address this problem, \cite{mao2015generation, yu2016refer, nagaraja16refexp, hu2015natural} model $P(r|o)$ and looks for the object $o$ maximizing the probability.
People also try modeling $P(o, r)$ directly using embedding model~\cite{rohrbach2015grounding, wang2015learning}, which learns to minimize the distance between paired object and sentence in the embedding space.
The generation task asks a system to compose an expression for a specified object within an image.
While some previous work used rule-based approaches to generate expressions with fixed grammar pattern~\cite{mitchell2013generating, fitzgerald2013learning, kazemzadeh2014referitgame}, recent work has followed the CNN-LSTM structure to generate expressions~\cite{mao2015generation, yu2016refer}. 

A Speaker and listener are typically considerred to formulate the two tasks, where speaker model is used for referring expression generation and listener is used for comprehension.
Golland et al~\cite{golland2010game} proposed that an optimal speaker should act based on listener's response.
Mao et al~\cite{mao2015generation} followed this idea by suppressing the ambiguity of listener in modeling $P(r|o)$.
Most related work to us is~\cite{andreas2016reasoning}, which uses a speaker model to generate expressions for abstract images then uses an offline listener to rerank the speaker's output. 
Compared with above, our model jointly trains the speaker and listener modules and has an additional reinforcer module to encourage unambiguous speaker expression generation. 
Moreover, we use our models for both generation and comprehension tasks on three natural image datasets.

\vspace{-.2cm}
\section{Model}
\label{sec:model}
\vspace{-.1cm}

Our model is composed of three modules: speaker (Sec~\ref{sec:speaker}), listener (Sec~\ref{sec:listener}), and reinforcer (Sec~\ref{sec:reinforcer}).
During training, the speaker and listener are trained jointly so that they can benefit from each other and from the reinforcer.  Because the reward function for the reinforcer is not differentiable, it is incorporated during training using a reinforcement learning policy gradient algorithm.

\vspace{-.1cm}
\subsection{Speaker}
\label{sec:speaker}
\vspace{-.1cm}

For our speaker module, 
we follow the previous state-of-the-art~\cite{mao2015generation, yu2016refer}, and use a CNN-LSTM framework.  Here, a pre-trained CNN model is used to define a visual representation for the target object and other visual context. Then, a Long-short term memory (LSTM) is used to generate the most likely expression given the visual representation. 

Because of the improved quantitative performance over~\cite{mao2015generation}, we use the visual comparison model of ~\cite{yu2016refer} as our speaker (to encode the target object). Here, the visual representation includes the target object, context, location/size features, and two visual comparison features.
Specifically, the target object representation $o_i$ is modeled as the fc7 features from a pre-trained VGG network~\cite{simonyan2014very}. 
Global context, $g_i$, is modeled as features extracted from the VGG-fc7 layer for the entire image.
The location/size representation, $l_i$, for the target object is modeled as a 5-dimensional vector, encoding the x and y locations of the top left and bottom right corners of the target object bounding box, as well as the bounding box size with respect to the image, i.e., $l_i=[\frac{x_{tl}}{W}, \frac{y_{tl}}{H}, \frac{x_{br}}{W}, \frac{y_{br}}{H}, \frac{w\cdot h}{W\cdot H}]$.

As referring expressions often relate an object to other objects of the same 
type within the image (``the red ball'' vs ``the blue ball'' or ``the larger elephant''), comparisons tend to be quite important for differentiation. The comparison features are composed of two parts: a) appearance similarity -- $\delta v_i=\frac{1}{n}\sum_{j\neq i} \frac{o_i - o_j}{\| o_i - o_j \|}$, where $n$ is the number of objects chosen for comparisons, b) location and size similarity -- $\delta l_i$, concatenating the 5-d difference on each compared object $\delta l_{ij}=[\frac{[\bigtriangleup x_{tl}]_{ij}}{w_i}, \frac{[\bigtriangleup y_{tl}]_{ij}}{h_i}, \frac{[\bigtriangleup x_{br}]_{ij}}{w_i}, \frac{[\bigtriangleup y_{br}]_{ij}}{h_i}, \frac{w_j h_j}{w_i h_i}]$. 

The final visual representation for the target object is then a concatenation of the above features followed by a fully-connected layer fusing them together, $r_i = W_m [o_i, g_i, l_i, \delta v_i, \delta l_i] + b_m$.
This joint feature is then fed into the LSTM for referring expression generation.
During training we minimize the negative log-likelihood: 

\begin{align}
\begin{split}
L_1^s(\theta) & = -\sum_i \log P(r_i|o_i;\theta) \\
              & = -\sum_i \sum_t \log P(r^t_i|r^{t-1}_i, \ldots, r^1_i, o_i;\theta)
\end{split}
\end{align}
Note that the speaker can be modeled using any form of CNN-LSTM structure.

In~\cite{mao2015generation}, Mao proposed to add a Maximum Mutual Information (MMI) constraint encouraging the generated expression to describe the target object better than the other objects within the image (i.e., a ranking loss on objects).
We generalize this idea to incorporate two triplet hinge losses composed of a positive match and two negative matches. 
Given a positive match $(r_i, o_i)$, we sample the contrastive pair $(r_j, o_i)$ where $r_j$ is the expression describing some other object and pair $(r_i, o_k)$ where $o_k$ is some other object in the same image, then we optimize the following max-margin loss:
\begin{equation}
\begin{split}
L_2^s(\theta) = \sum_i [ \lambda^s_1 \max(0, M+\log P(r_i|o_k)-\log P(r_i|o_i)) \\
+ \lambda^s_2 \max(0, M+\log P(r_j|o_i) - \log P(r_i|o_i))]
\end{split}\label{eqn:speaker}
\end{equation}
The first term is from~\cite{mao2015generation}, while the second term  encourages that the target object to be better described by the true expression compared to expressions describing other objects in the image (i.e., a ranking loss on expressions).

\vspace{-.1cm}
\subsection{Listener}
\label{sec:listener}
\vspace{-.1cm}

We use a joint-embedding model to mimick the listener's behaviour.
The purpose of this embedding model is to encode the visual information from the target object and semantic information from the referring expression into a joint embedding space that embeds vectors that are visually or semantically related closer together in the space. 
Here for referring expression comprehension task, given a referring expression representation, the listener embeds it into the joint space, then selects the closest object in the embedding space for the predicted target object.

As illustrated in Fig.~\ref{fig:model}, for our listener joint-embedding model (outlined by a red dashline), we use an LSTM to encode the input referring expression and the same visual representation as the speaker to encode the target object (thus connecting the speaker to the listener).
We then add two MLPs (multi-layer perceptions) and two L2 normalization layers following each view, the object and the expression.
Each MLP is composed of two fully connected layers with ReLU nonlinearities between them, serving to transform the object view and the expression view into a common embedding space.
The inner-product of the two normalized representations is computed as their similarity score $S(r, o)$ in the space.
As a listener, we force the similarity on target object and referring expression pairs by applying a hinge loss over triplets, which consist of a positive match and two negative matches:
\begin{equation}
\begin{split}
L^l(\theta) = \sum_i [ \lambda^l_1 \max(0, M+S(r_i, o_k)-S(r_i, o_i)) \\
+ \lambda^l_2 \max(0, M+S(r_j, o_i) - S(r_i, o_i))]
\end{split}\label{eqn:listener}
\end{equation}
where the negative matches are randomly chosen from the other objects and expressions in the same image.

Note that the listener model is not limited to this particular triplet-based model. For example, \cite{rohrbach2015grounding} computes a similarity score between every object for given referring expression, and minimizes the cross entropy of the SoftMax knowing the target object, which could also be applied here.

\vspace{-.1cm}
\subsection{Reinforcer}
\label{sec:reinforcer}
\vspace{-.1cm}

Besides using the ground-truth pairs of target object and referring expression for training the speaker, we also use reinforcement learning to guide the speaker toward generating less ambiguous expressions (expressions that apply to the target object but not to other objects).
This reinforcer module is composed of a discriminative reward function and performs a non-differentiable policy gradient update to the speaker.

Specifically, given the softmax output of the speaker's LSTM, we sample words according to the categorical distribution at each time step, resulting in a complete expression after sampling the $<$END$>$ token.
This sampling operation is non-differentiable as we do not know whether an expression is ambiguous or not until we feed it into a reward function.
Therefore, we use policy gradient reinforcement learning to update the speaker's parameters.
Here, the goal is to maximize the reward expectation $F(w_{1:T})$ under the distribution of $p(w_{1:T}; \theta)$ parameterized by the speaker, i.e., $J = E_{p(w_{1:T})}[F]$.
According to the policy gradient algorithm~\cite{williams1992simple}, we have
\begin{equation}
\nabla_{\theta}J = E_{p(w_{1:T})}[F(w_{1:T})\nabla_{\theta}\log p(w_{1:T};\theta)],\label{eqn:policy}
\end{equation}
Where $\log p(w_t)$ is defined by the softmax output.
We then use this gradient to update our speaker model during training.

The only thing left is to choose a reward function that encourages the speaker to sample less ambiguous expressions.
As illustrated in Fig.~\ref{fig:model} (outlined in dashed orange), the reinforcer module learns a reward function using paired objects and expressions.
We again use the same visual representation for the target object and use another LSTM to encode the expression representation.
Rather than using two MLPs to encode each view as in the listener, here we concatenate the two views and feed them together into a MLP to learn a 1-d Logistic Regression score between 0 and 1.
Trained with cross-entropy loss, the reward function computes a match score between an input object and expression.
We use this score as the reward signal in Eqn.~\ref{eqn:policy} for sampled expression and target object pairs.
After training, the reward function is fixed to assist our joint speaker-listener system.

\vspace{-.1cm}
\subsection{Joint Model}
\vspace{-.1cm}

In this subsection, we describe some specifics of how our three modules (speaker, listener, reinforcer) are integrated into a joint framework (shown in Fig.~\ref{fig:model}).
For the listener, we notice that the visual vector in the embedding space is learned to capture the neighbourhood vectors of referring expressions, thus making it aware of the listener's knowledge. 
Therefore, we take this MLP embedded vector as an additional input for the speaker, which encodes the listener based information.
In Fig.~\ref{fig:model}, we use concatenation to jointly encode the standard visual representation of target object and this listener-aware representation and then feed them into speaker.
Besides concatenation, the element-wise product or compact bilinear pooling can also be applied~\cite{fukui2016multimodal}.
During training, we sample the same triplets for both the speaker and listener, and make the word embedding of the speaker and listener shared to reduce the number of parameters.
For the reinforcer module, we do sentence sampling using the speaker's LSTM as shown in the top right of Fig.~\ref{fig:model}.
Within each mini-batch, the sampled expressions for the target objects are fed into the reward function to obtain reward values.

The overall loss function is formulated as a multi-task learning problem:
\begin{equation}
\theta = \arg\min L_1^s(\theta) + L_2^s(\theta) + L^l(\theta) - \lambda^r J(\theta),
\end{equation}
where $\lambda^r$ is the weight on reward loss.
The weights on the loss of speaker and listener are already included in Eqn.~\ref{eqn:speaker} and Eqn.~\ref{eqn:listener}.
We list all hyper-parameters settings in Sec.~\ref{optimization}.

\vspace{-.1cm}
\subsection{Comprehension and Generation}
\vspace{-.1cm}
For the comprehension task, 
at test time, we could use either the speaker or listener to select the target object given an input expression. Using the listener, we would embed the input expression into the learned embedding space and select the closest object as the predicted target. Using the speaker, we would generate expressions for each object within the image and then select the object whose generated expression best matches the input expression.
Therefore, we utilize both modules by ensembling the speaker and listener predictions together to pick the most probable object given a referring expression.

\begin{equation}
\hat{o} = \arg\max_o P(r|o) S(o, r)^{\lambda}
\end{equation}

Surprisingly, using the speaker alone (setting $\lambda$ to 0) already achieves state-of-art results due to the joint training.
Adding the listener further improves performance to more than 4\% over previous state-of-art results.

For the generation task, we first let the speaker generate multiple expressions per object via beam search.
We then use the listener to rerank these expressions and select the least ambiguous expression, which is similar to~\cite{andreas2016reasoning}.
To fully utilize the listener's power in generation, we propose to consider cross comprehension as well as the diversity of expressions by minimizing the potential:

\begin{align}
\begin{split}
&E(r) = \sum_i \theta_i (r_i) + \sum_{i,j} \theta_{i,j}(r_i, r_j) \\
&\theta_i(r_i)  = - \log P (r_i|o_i) - \lambda_1 \log S(r_i, o_i) \\
& \qquad \qquad  + \lambda_2 \max_{j\neq i} \log S(r_i, o_j)\\
&\theta_{i,j}(r_i, r_j)  = \lambda_3 I(r_i = r_j)
\end{split}\label{eqn:generation}
\end{align} 

The first term and second term in unary potential measure how well the target object and generated expression match using the speaker and listener modules respectively, which was also used in~\cite{andreas2016reasoning}.
The third term in unary potential measures the likelihood of the generated sentence for describing other objects in the same image. 
The pairwise potential penalize the same sentences being generated for different objects (encouraging diversity in generation).
In this way, the expressions for every object in an image are jointly generated.
Compared with the previous model that attempted to tie language generation of referring expressions together~\cite{yu2016refer}, the constraints in Eqn.~\ref{eqn:generation} are more explicit and overall this works better to reduce ambiguity in the generated expressions.

\begin{table*}[t]
\footnotesize
\begin{center}
\resizebox{2.0\columnwidth}{!}{%
\begin{tabular}{| c |  l | c | c | c || c | c | c || c |}
\hline
&& \multicolumn{3}{c}{RefCOCO} & \multicolumn{3}{|c|}{RefCOCO+} & RefCOCOg\\
\cline{2-9}
&& val & TestA & TestB & val & TestA & TestB & val\\
\hline\hline
1&listener 					 			& 77.48\% & 76.58\% & 78.94\% & 60.50\% & 61.39\% & 58.11\% & 71.12\% \\
2&previous state-of-art\cite{nagaraja16refexp}\cite{yu2016refer} & 76.90\% & 75.60\% & 78.00\%\cite{nagaraja16refexp} & 58.94\% & 61.29\% & 56.24\%\cite{yu2016refer} & 65.32\%\cite{yu2016refer} \\
\hline
3&baseline\cite{mao2015generation} 	 	& 64.56\% & 63.20\% & 66.69\% & 47.78\% & 51.01\% & 44.24\% & 56.81\% \\
4&\textbf{speaker}\cite{yu2016refer} 		& 69.95\% & 68.59\% & 72.84\% & 52.63\% & 54.51\% & 50.02\% & 59.40\% \\
5&\textbf{speaker}+listener				& 71.20\% & 69.98\% & 73.66\% & 54.23\% & 56.22\% & 52.46\% & 61.83\% \\
6&\textbf{speaker}+reinforcer 				& 71.88\% & 70.18\% & 73.01\% & 53.38\% & 56.50\% & 51.16\% & 61.91\% \\
7&\textbf{speaker}+listener+reinforcer 	& 72.46\% & 71.10\% & 74.01\% & 55.54\% & 57.46\% & 53.71\% & 64.07\% \\
\hline
8&baseline+MMI\cite{mao2015generation} 	& 72.28\% & 72.60\% & 73.39\% & 56.66\% & 60.01\% & 53.15\% & 63.31\% \\
9&\textbf{speaker}+MMI\cite{yu2016refer}  & 76.18\% & 74.39\% & 77.30\% & 58.94\% & 61.29\% & 56.24\% & 65.32\% \\
10&\textbf{speaker}+listener+MMI 			& 79.22\% & 77.78\% & 79.90\% & 61.72\% & 64.41\% & 58.62\% & 71.77\% \\
11&\textbf{speaker}+reinforcer+MMI 			& 78.38\% & 77.13\% & 79.53\% & 61.32\% & 63.99\% & 58.25\% & 67.06\% \\
12&\textbf{speaker}+listener+reinforcer+MMI & {\bf 79.56\%} & {\bf 78.95\%} & {\bf 80.22\%} & {\bf 62.26\%} & {\bf 64.60\%} & {\bf 59.62\%} & {\bf 72.63\%} \\
\hline
\hline
&& \multicolumn{3}{c}{RefCOCO (detected)} & \multicolumn{3}{|c|}{RefCOCO+ (detected)} & RefCOCOg (detected)\\
\cline{2-9}
&& val & TestA & TestB & val & TestA & TestB & val\\
\hline\hline
1&listener					 			& - & 71.63\% & 61.47\% & - & 57.33\% & 47.21\% & 56.18\% \\
2&previous state-of-art\cite{yu2016refer} & - & 72.03\% & 63.08\% & - & 58.87\% & 47.70\% & 58.26\% \\
\hline
3&baseline\cite{mao2015generation} 	 	& - & 64.42\% & 56.75\% & - & 52.84\% & 42.68\% & 53.13\% \\
4&\textbf{speaker}\cite{yu2016refer} 		& - & 67.69\% & 60.16\% & - & 54.37\% & 45.00\% & 53.83\% \\
5&\textbf{speaker}+listener			 	& - & 68.27\% & 61.00\% & - & 55.41\% & 45.65\% & 54.96\% \\
6&\textbf{speaker}+reinforcer 			 	& - & 69.12\% & 60.47\% & - & 55.45\% & 44.96\% & 55.64\% \\
7&\textbf{speaker}+listener+reinforcer     & - & 69.15\% & 61.96\% & - & 55.97\% & 46.45\% & 57.03\% \\
\hline
8&baseline+MMI\cite{mao2015generation} 	& - & 68.73\% & 59.56\% & - & 58.15\% & 46.63\% & 57.23\% \\
9&\textbf{speaker}+MMI\cite{yu2016refer}  & - & 72.03\% & 63.08\% & - & 58.87\% & 47.70\% & 58.26\% \\
10&\textbf{speaker}+listener+MMI 			& - & {\bf 72.95\%} & 63.10\% & - & 60.23\% & 48.11\% & 58.57\%\\
11&\textbf{speaker}+reinforcer+MMI 		& - & 72.34\% & 63.24\% & - & 59.36\% & 48.72\% & 58.70\% \\
12&\textbf{speaker}+listener+reinforcer+MMI& - & 72.88\% & {\bf 63.43\%} & - & {\bf 60.43\%} & {\bf 48.74\%} & {\bf 59.51\%} \\
\hline
\end{tabular}
}
\end{center}
\vspace{-10pt}
\caption{Ablation study using the speaker module for the comprehension task (indicated in {\bf bold}). Top half shows performance given ground truth bounding boxes for objects, bottom half performance using automatic object detectors to select potential objects. We find that adding listener and reinforcer modules to the speaker increases performance.}
\label{table:comprehension_speaker}
\end{table*}

\begin{table*}[t]
\footnotesize
\begin{center}
\resizebox{2.0\columnwidth}{!}{%
\begin{tabular}{| c | l | c | c | c || c | c | c || c |}
\hline
&& \multicolumn{3}{c}{RefCOCO} & \multicolumn{3}{|c|}{RefCOCO+} & RefCOCOg\\
\cline{2-9}
&& val & TestA & TestB & val & TestA & TestB & val\\
\hline\hline
1&listener									      & 77.48\% & 76.58\% & 78.94\% & 60.50\% & 61.39\% & 58.11\% & 71.12\%\\
2&previous state-of-art~\cite{nagaraja16refexp}\cite{yu2016refer} & 76.90\% & 75.60\% & 78.00\%~\cite{nagaraja16refexp} & 58.94\% & 61.29\% & 56.24\%~\cite{yu2016refer} & 65.32\%~\cite{yu2016refer} \\
\hline
3&speaker+\textbf{listener} 					   & 77.84\% & 77.50\% & 79.31\% & 60.97\% & 62.85\% & 58.58\% & 72.25\% \\
4&speaker+\textbf{listener}+reinforcer 			   & 78.14\% & 76.91\% & 80.10\% & 61.34\% & 63.34\% & 58.42\% & 71.72\%\\
5&\textbf{speaker+listener}+reinforcer (ensemble)  & 78.88\% & 78.01\% & 80.65\% & 61.90\% & 64.02\% & 59.19\% & 72.43\% \\
\hline
6&speaker+\textbf{listener}+MMI 				   & 78.42\% & 78.45\% & 79.94\% & 61.48\% & 62.14\% & 58.91\% & 72.13\% \\
7&speaker+\textbf{listener}+reinforcer+MMI		   & 78.36\% & 77.97\% & 79.86\% & 61.33\% & 63.10\% & 58.19\% & 72.02\% \\
8&\textbf{speaker+listener}+reinforcer+MMI (ensemble)& {\bf 80.36\%} & {\bf 80.08\%} & {\bf 81.73\%} & {\bf 63.83\%} & {\bf 65.40\%} & {\bf 60.73\%} & {\bf 74.19\%} \\
\hline
\hline
&& \multicolumn{3}{c}{RefCOCO (detected)} & \multicolumn{3}{|c|}{RefCOCO+ (detected)} & RefCOCOg (detected)\\
\cline{2-9}
&& val & TestA & TestB & val & TestA & TestB & val\\
\hline\hline
1&listener					 			  		   & - & 71.63\% & 61.47\% & - & 57.33\% & 47.21\% & 56.18\% \\
2&previous state-of-art\cite{yu2016refer} 		   & - & 72.03\% & 63.08\% & - & 58.87\% & 47.70\% & 58.26\% \\
\hline
3&speaker+\textbf{listener} 					   & - & 72.23\% & 62.92\% & - & 59.61\% & 48.31\% & 57.38\% \\
4&speaker+\textbf{listener}+reinforcer 			   & - & 72.65\% & 62.69\% & - & 58.68\% & 48.23\% & 58.32\% \\
5&\textbf{speaker+listener}+reinforcer (ensemble)  & - & 72.78\% & {\bf 64.38\%} & - & 59.80\% & 49.34\% & {\bf 60.46\%} \\
\hline
6&speaker+\textbf{listener}+MMI 				   & - & 72.95\% & 62.43\% & - & 58.68\% & 48.44\% & 57.34\% \\
7&speaker+\textbf{listener}+reinforcer+MMI		   & - & 72.94\% & 62.98\% & - & 58.68\% & 47.68\% & 57.72\% \\
8&\textbf{speaker+listener}+reinforcer+MMI (ensemble)& - & {\bf 73.78\%} & 63.83\% & - & {\bf 60.48\%} & {\bf 49.36\%} & 59.84\% \\
\hline
\end{tabular}
}
\end{center}
\vspace{-10pt}
\caption{Ablation study using listener or ensembled listener+speaker modules for the comprehension task (indicated in {\bf bold}). Top half shows performance given ground truth bounding boxes for objects, bottom half performance using automatic object detectors to select potential objects. We find that jointly training with the speaker improves listener's performance and that adding the reinforcer module in an ensembled speaker+listener prediction performs the best.}
\label{table:comprehension_listener}
\end{table*}

\begin{figure*}[t]
\centering
\includegraphics[width=0.9\textwidth]{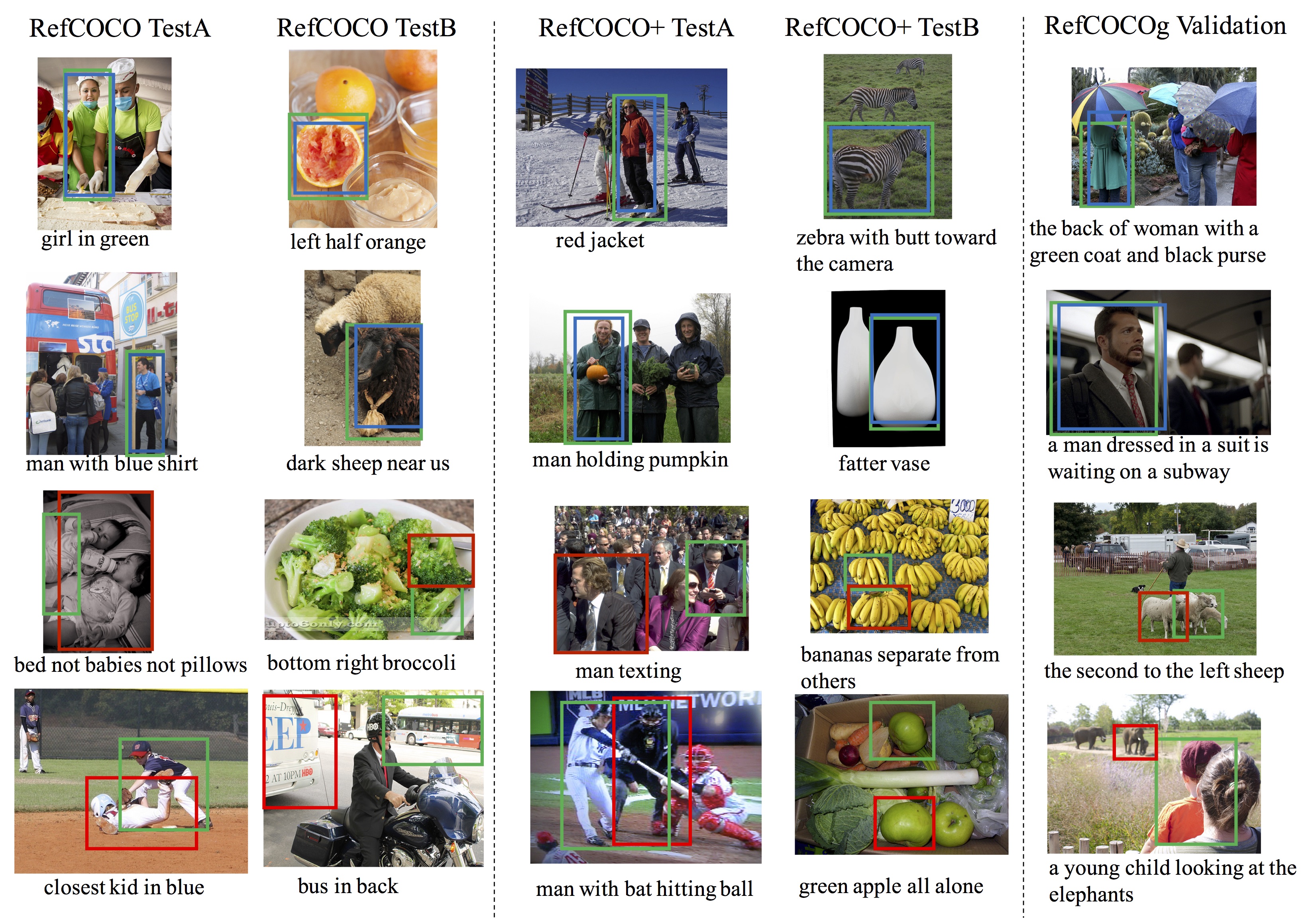}
\caption{Example comprehension results based on detection. Green box shows the ground-truth region, blue box shows correct comprehension using our ``speaker+listener+reinforcer+MMI'' model, and red box shows incorrect comprehension. We use top two rows to show some correct comprehensions and bottom two rows to show some incorrect ones.}
\label{fig:comprehension}
\end{figure*}

\section{Experiments}

\subsection{Optimization}\label{optimization}
We optimize our model using Adam~\cite{kingma2014adam} with an initial learning rate of 0.0004, halved every 2,000 iterations, with a batch size of 32.
The word embedding size and hidden state size of the LSTM are set to 512.
We also set the output of each MLP and FC in Fig.~\ref{fig:model} to be 512-dimensional.
To avoid overfitting, we apply dropout with a ratio of 0.2 after each linear transformation in the MLP layers of the listener.
We also regularize the word-embedding and output layers of the LSTM in the speaker using dropout with ratio of 0.5.
For the constrastive pairs, we set $\lambda_1^l=1$ and $\lambda_2^l=1$ in the listener's objective function (Eqn.~\ref{eqn:listener}), and set $\lambda_1^s = 1$ and $\lambda_2^s=0.1$ in the speaker's objective function (Eqn.~\ref{eqn:speaker}).
We emprically set the weight on reward loss $\lambda^r=1$ .

\begin{table*}[t]
\footnotesize
\begin{center}
\resizebox{2\columnwidth}{!}{%
\begin{tabular}{| l | c | c | c | c | c | c | c | c | c | c |}
\hline
& \multicolumn{4}{c|}{RefCOCO} & \multicolumn{4}{c|}{RefCOCO+} & \multicolumn{2}{c|}{RefCOCOg}\\
\cline{2-11}
&  \multicolumn{2}{c|}{Test A} & \multicolumn{2}{c|}{Test B} &  \multicolumn{2}{c|}{Test A} & \multicolumn{2}{c|}{Test B} & \multicolumn{2}{c|}{val}\\
\cline{2-11}
& Meteor & CIDEr & Meteor & CIDEr & Meteor & CIDEr & Meteor & CIDEr & Meteor & CIDEr \\
\hline\hline
speaker+tie~\cite{yu2016refer} & 0.283 & 0.681 & 0.320 & 1.273 & 0.204 & 0.499 & 0.196 & 0.683 & - & - \\ 
\hline
baseline+MMI      						& 0.243 & 0.615 & 0.300 & 1.227 & 0.199 & 0.462 & 0.189 & 0.679 & 0.149 & 0.585\\
speaker+MMI 							& 0.260 & 0.679 & 0.319 & 1.276 & 0.202 & 0.475 & 0.196 & 0.683 & 0.147 & 0.573\\
speaker+listener+MMI 					& 0.268 & 0.704 & 0.327 & 1.303 & \bf{0.208} & \bf{0.496} & 0.201 & 0.697 & 0.150 & 0.589\\
speaker+reinforcer+MMI					& 0.266 & 0.702 & 0.323 & 1.291 & 0.204 & 0.482 & 0.197 & 0.692 & 0.151 & \bf{0.602}\\
speaker+listener+reinforcer+MMI 		& \bf{0.268} & 0.697 & \bf{0.329} & \bf{1.323} & 0.204 & 0.494 & \bf{0.202} & \bf{0.709} & \bf{0.154} & 0.592\\
\hline
\hline
baseline+MMI+rerank      				& 0.280 & 0.729 & 0.329 & 1.285 & 0.204 & 0.484 & 0.205 & 0.730 & \bf{0.160} & 0.654\\
speaker+MMI+rerank 						& 0.287 & 0.745 & 0.334 & 1.295 & 0.208 & 0.490 & 0.213 & 0.712 & 0.156 & 0.653\\
speaker+listener+MMI+rerank 			& 0.293 & 0.763 & 0.337 & 1.306 & 0.211 & 0.500 & \bf{0.221} & 0.734 & 0.159 & 0.650\\
speaker+reinforcer+MMI+rerank			& 0.291 & 0.748 & 0.337 & 1.311 & 0.207 & 0.499 & 0.215 & 0.729 & 0.158 & 0.653\\
speaker+listener+reinforcer+MMI+rerank 	& \bf{0.296} & \bf{0.775} & \bf{0.340} & \bf{1.320} & \bf{0.213} & \bf{0.520} & 0.215 & \bf{0.735} & 0.159 & \bf{0.662}\\
\hline
\end{tabular}
}
\end{center}
\caption{Ablation study for generation using automatic evaluation.\vspace{-7pt}}
\label{table:metric}
\end{table*}

\subsection{Datasets}
We perform experiments on three referring expression datasets: RefCOCO, RefCOCO+ and RefCOCOg (described in Sec~\ref{sec:relatedwork}).
All three datasets are collected on MSCOCO images~\cite{lin2014microsoft}, but with several differences: 
1) RefCOCO and RefCOCO+ were collected using an interactive game interface 
while RefCOCOg was collected in a non-interactive setting and contains longer expressions,
2) RefCOCOg contains on average 1.63 objects of the same type per images,
while RefCOCO and RefCOCO have 3.9 on average, 
3) RefCOCO+ disallowed absolute location words in referring expressions.
Overall, RefCOCO has 142,210 expressions for 50,000 objects in 19,994 images, RefCOCO+ has 141,565 expressions for 49,856 objects in 19,992 images, and RefCOCOg has 104,560 expressions for 54,822 objects in 26,711 images.

Additionally, each dataset is provided with dataset splits for evaluation.
RefCOCO and RefCOCO+ provide person vs. object splits for evaluation. Images containing multiple people are in ``TestA'' while images containing multiple objects of other categories are in ``TestB''.
For RefCOCOg, the authors divide their dataset by randomly partitioning objects into training and testing splits. Thus same image may appear in the two splits.
As only training and validation splits have been released for this dataset, we use the hyper-paramters cross-validated on RefCOCO to train models on RefCOCOg.

\subsection{Comprehension Task}
After training, we can either use the speaker or listener to perform the comprehension task.
For the speaker that models $P(r|o)$, we feed every ground-truth object region within the given image to the speaker and select the most probable region for the expression as the comprehension result, i.e., $o^{*}=\mbox{argmax}_{o_i} p(r|o_i)$.
For the listener, we directly compute the similarity score $S(r, o)$ between the proposal/object and expression and pick the object with the highest probability. 
For evaluation, we compute the intersection-over-union (IoU) of the comprehended region with the ground-truth object.
If the IoU score of the predicted region is greater than 0.5, we consider this a correct comprehension.

To demonstrate the benefits of each module, we run ablation studies in Lines 3-12 of Table~\ref{table:comprehension_speaker} (for speaker as comprehender) and in Lines 3-8 of Table~\ref{table:comprehension_listener} (for listener as comprehender) on all three datasets. 
Each row shows the results after adding a module during training.
For some models that have both speaker and listener, we highlight the module being used for comprehension in bold.
For example, ``\textbf{speaker}+listener'' means we use the speaker module of the joint model to do the comprehension task, while ``speaker+\textbf{listener}'' means we use the listener module for this task.
Note our speaker module is implemented using the ``visdif'' model in~\cite{yu2016refer} as mentioned in Section~\ref{sec:speaker}\footnote{The ``visdif'' model trained in this paper performs slightly better on comprehension task than the original one reported in~\cite{yu2016refer}.}.
Following previous work~\cite{mao2015generation}\cite{yu2016refer}\cite{nagaraja16refexp}, we show the results trained with MMI and those trained w/o MMI on speakers.
We compare our models with the ``baseline'' model~\cite{mao2015generation} (Line 3, 8 in Table~\ref{table:comprehension_speaker}), the pure listener model (Line 1), and previous state-of-art results (Line 2) achieved in~\cite{nagaraja16refexp}\cite{yu2016refer}.

First, we evaluate the performance of the speaker on the comprehension task (Table~\ref{table:comprehension_speaker}).
We observe all speaker models trained with MMI outperform w/o MMI.
We also find that the speaker can be improved by joint training with the listener module and by incorporating the reinforcer module.
With MMI ranking, the speaker learned with joint training (Line 12) is able to outperform the pure listener by around 2\% on all three datasets, which already achieves state-of-art performance on the comprehension task.

Second, we show evaluations using variations of the listener module or ensembled listener+speaker modules (indicated in {\bf bold}) for the comprehension task in Table~\ref{table:comprehension_listener}.
We note that the listener generally works better than speaker for the comprehension task, indicating that the deterministic joint-embedding model is more suitable for this task than the speaker model -- similar results were observed in~\cite{rohrbach2015grounding}.
While the reinforcer module seems not to be as effective as the speaker, we still find that the joint training always brings additional discriminative benefits to the listener module, resulting in improved performance (compare Line 3-8 with Line 1 in Table~\ref{table:comprehension_listener}).
Ensembling the speaker and listener together achieves the best results overall.

Both of the above experiments analyze comprehension performance given ground truth bounding boxes for potential comprehension objects, where the algorithm must select which of these objects is being described. 
This provides an analysis of comprehension performance independent of any particular object detection method.
Additionally, we also show results using an object detector to  automatically select regions for consideration during comprehension in the bottom half of each table (Tables~\ref{table:comprehension_speaker} and~\ref{table:comprehension_listener}).
As our detection algorithm, we use current state of the art detector in effectiveness and speed, SSD~\cite{liu2015ssd}, trained on a subset of the MS COCO train+val dataset, removing images that are in the test splits of RefCOCO or RefCOCO+ or in the validation split of RefCOCOg.
We empirically select 0.30 as the confidence threshold for detection output.
While performance drops somewhat due to the strong dependence of ``visdif'' model on detection~\cite{yu2016refer}, the overall improvements brought by each module are consistent with using ground-truth objects, showin the robustness of our joint model.
Fig.~\ref{fig:comprehension} shows some comprehension results using our full model.

\begin{table}[t]
\scriptsize
\centering
\resizebox{\columnwidth}{!}{%
\begin{tabular}{| l | c | c | c | c |}
\hline
& \multicolumn{2}{c}{RefCOCO} & \multicolumn{2}{c|}{RefCOCO+} \\
\cline{2-5}
&\ \ Test A\ \  &\ \ Test B\ \   &\ \ Test A\ \  &\ Test B\ \\
\hline\hline
speaker+tie\cite{yu2016refer}& 71.40\% & 76.14\% & 57.17\% & 47.92\% \\
\hline
speaker+MMI~\cite{yu2016refer} 			& 68.82\% & 75.50\% & 53.57\% & 47.88\% \\
speaker+listener+MMI 					& 73.23\% & 76.08\% & 53.83\% & 49.19\% \\
speaker+reinforcer+MMI 					& 71.08\% & 76.09\% & 55.16\% & 48.50\% \\
speaker+listener+reinforcer+MMI 		& 74.08\% & 76.44\% & 56.92\% & 53.23\% \\
\hline
speaker+listener+reinforcer+MMI& & & & \\+rerank   & \bf{76.95\%} & \bf{78.10\%} & \bf{58.85\%} & \bf{58.20\%} \\
\hline
\end{tabular}
}
\caption{Human Evaluations on generation.}
\label{table:human_eval}
\end{table}

\subsection{Generation Task}
\vspace{-.1cm}
For the generation task, we evaluate variations on the speaker module. 
Evaluating the generation is not as simple as comprehension.
In image captioning, BLEU, ROUGE, METEOR and CIDEr are common automatic metrics and have been widely used as standard evaluations.
We show automatic evaluation using the METEOR and CIDEr metrics for generation in Table~\ref{table:metric} where ``+rerank'' denotes models incorporating the reranking mechanism and global optimization over all objects (Eqn.~\ref{eqn:generation}).  
To computer CIDEr robustly, we collect more expressions for objects in the test sets for RefCOCO and RefCOCO+, obtaining 10.1 and 9.4 expressions respectively on average per object.
For RefCOCOg we use the original expressions released with the dataset which may be limited, but we still show its performance for completeness.
We choose the ``speaker+tie'' model in~\cite{yu2016refer} as reference, which learns to tie the expression generation together and achieves state-of-art performance.
Generally we find that the speaker in jointly learned models achieves higher scores than the single speaker under both metrics across datasets.
Such improvements are observed under both settings without ``+rerank'' or with ``+rerank''.

In addition, since previous work~\cite{yu2016refer} has found that these metrics do not always agree well with human judgments for referring expressions, we also run a human evaluation on the same set of objects as~\cite{yu2016refer} for RefCOCO and RefCOCO+.  
Here we ask Turkers to click on the referred object given a generated expression.
These results are shown in Table.~\ref{table:human_eval}. 
Results indicate the ablated benefits brought by each module, and ultimately the ``+rerank'' of our joint model achieves the best performance.

\begin{figure*}[t]
\centering
\includegraphics[width=0.95\textwidth]{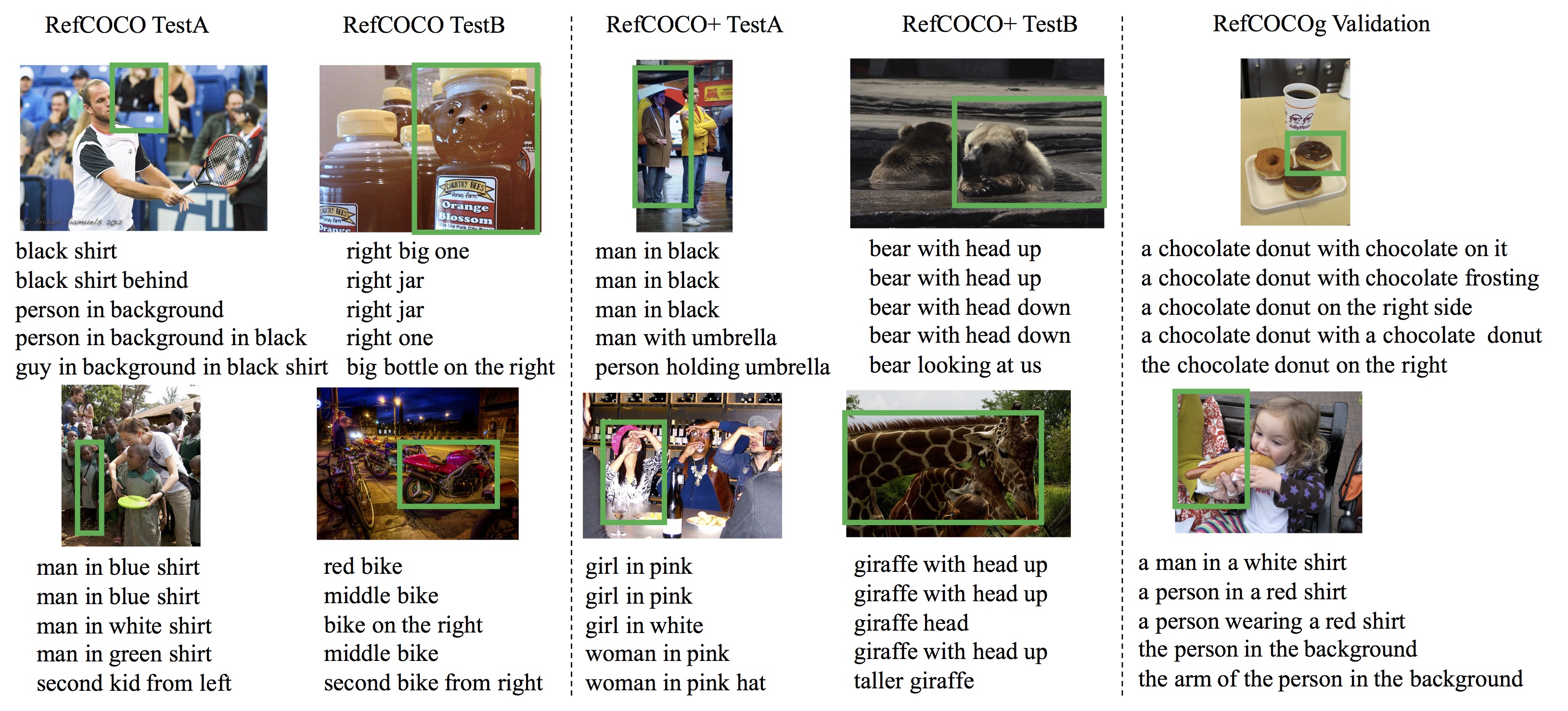}
\caption{Example generation results from each dataset. From top to bottom showing: speaker+MMI, speaker+listener+MMI, speaker+reinforcer+MMI, speaker+listener+reinforcer+MMI, speaker+listener+reinforcer+MMI+rerank.}
\label{fig:generation}
\end{figure*}

\begin{figure*}[t]
\centering
\includegraphics[width=0.9\textwidth]{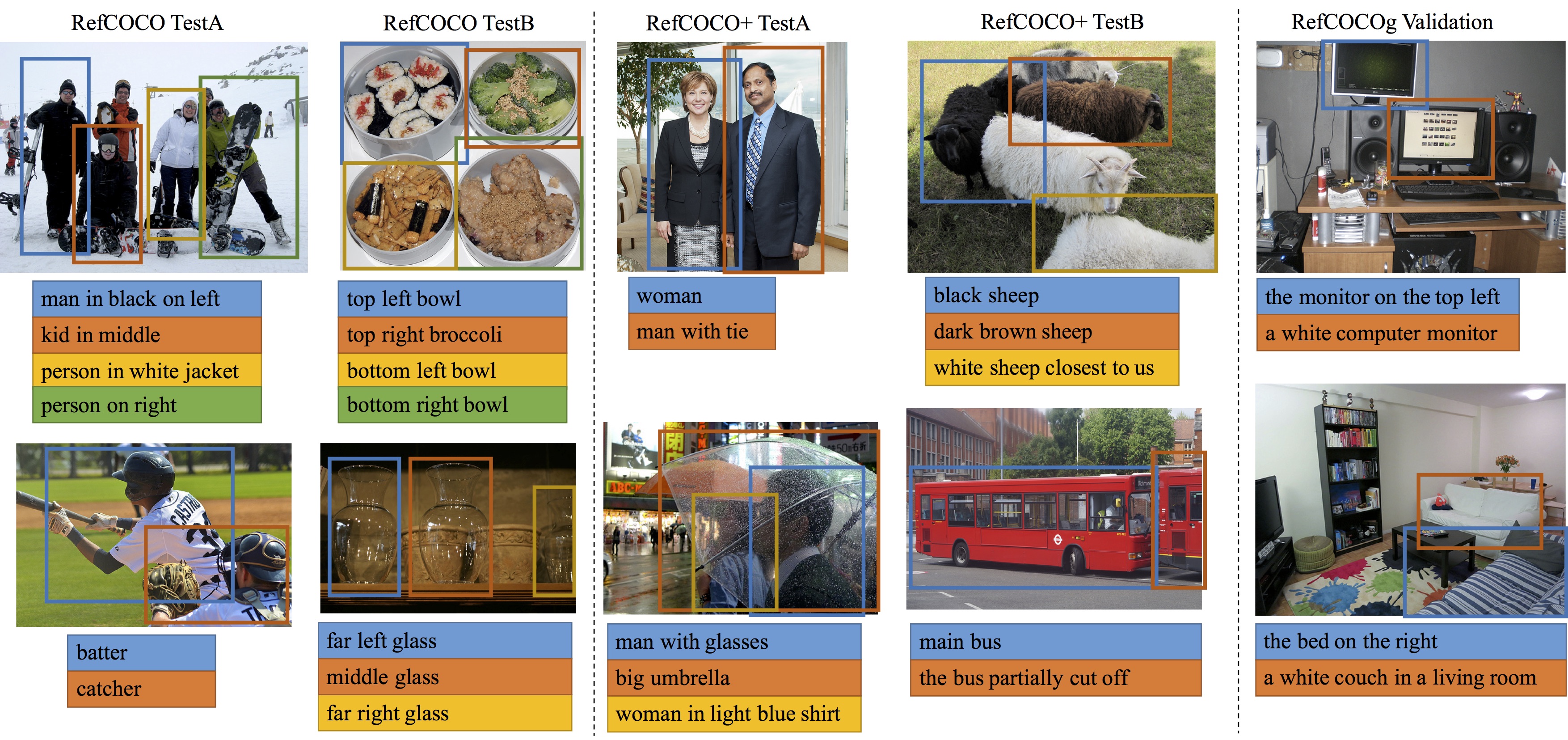}
\caption{Joint generation examples using ``speaker+listener+reinforcer+MMI+rerank''. Each sentence shows the generated expression for one of the depicted objects (color coded to indicate correspondence)}
\label{fig:joint_generation}
\end{figure*}

In Fig.~\ref{fig:generation}, we compare the generated expressions using the speaker module of different models.
We then show the joint expression generation using our full model with ``+rerank'' in Fig.~\ref{fig:joint_generation}.
As observed, the expressions of every target object are considered together, where each of them is meant to be relevant to the target object and irrelevant to the other objects.

\vspace{-.2cm}
\section{Conclusion}
\vspace{-.2cm}
We demonstrated the effectiveness of a unified framework for referring expression generation and comprehension. Our model consists of speaker and listener modules trained jointly to improve performance and a reinforcer module to help produce less ambiguous expressions. Experiments indicate that our model outperforms state of the art for both comprehension and generation on multiple datasets and evaluation metrics.

{\small
\bibliographystyle{ieee}
\bibliography{egbib}
}

\end{document}